\documentclass[journal,twoside,web]{ieeecolor}
\usepackage{tmi}
\usepackage{cite}
\usepackage{amsmath,amssymb,amsfonts}
\usepackage{algorithmic}
\usepackage{graphicx}
\usepackage{textcomp}
\newtheorem{problem}{Problem}
\def\BibTeX{{\rm B\kern-.05em{\sc i\kern-.025em b}\kern-.08em
    T\kern-.1667em\lower.7ex\hbox{E}\kern-.125emX}}
\begin{document}
\title{Physics-based Learning of Parameterized Thermodynamics from Real-time Thermography}
\author{Hamza El-Kebir, \IEEEmembership{Student Member, IEEE}, Yongseok Lee, and Joseph Bentsman, \IEEEmembership{Senior Member, IEEE}
\thanks{Manuscript submitted for review on July 19, 2022.
}
\thanks{H. El-Kebir is with the Dept. of Aerospace Engineering, University of Illinois Urbana-Champaign, Urbana, IL 61801 USA (e-mail: elkebir2@illinois.edu).}
\thanks{Y. Lee and J. Bentsman are with the Dept. of Mechanical Science and Engineering, University of Illinois Urbana- Champaign, Urbana, IL 61801 USA (email: \{yl50, jbentsma\}@illinois.edu). J.~Bentsman is the corresponding author.}}

\maketitle

\begin{abstract}
Progress in automatic control of thermal processes and real-time estimation of heat penetration into live tissue has long been limited by the difficulty of obtaining high-fidelity thermodynamic models. Traditionally, in complex thermodynamic systems, it is often infeasible to estimate the thermophysical parameters of spatiotemporally varying processes, forcing the adoption of model-free control architectures. This comes at the cost of losing any robustness guarantees, and implies a need for extensive real-life testing. In recent years, however, infrared cameras and other thermographic equipment have become readily applicable to these processes, allowing for a real-time, non-invasive means of sensing the thermal state of a process. In this work, we present a novel physics-based approach to learning a thermal process's dynamics \emph{directly} from such real-time thermographic data, while focusing attention on regions with high thermal activity. We call this process, which applies to any higher-dimensional scalar field, \emph{attention-based noise robust averaging} (ANRA). Given a partial-differential equation model structure, we show that our approach is robust against noise, and can be used to initialize optimization routines to further refine parameter estimates. We demonstrate our method on several simulation examples, as well as by applying it to electrosurgical thermal response data on \emph{in vivo} porcine skin tissue.
\end{abstract}


\begin{figure*}[htb]
\centering
\centerline{\includegraphics[width=\linewidth]{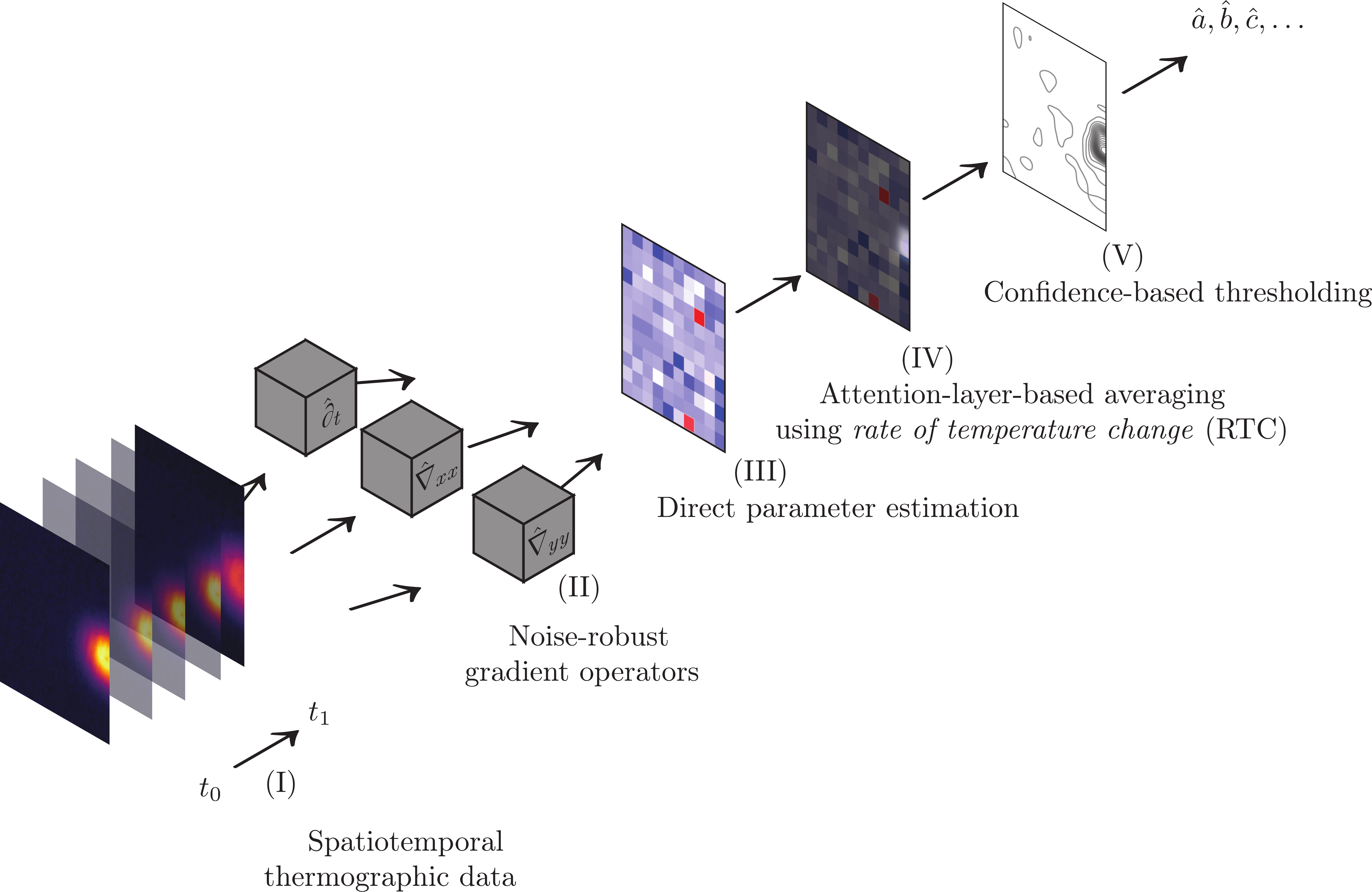}}
  \caption{Overall \emph{attention-based noise robust averaging} (ANRA) model architecture: (I) real-time thermographic imagery is acquired; (II) gradients that appear in the PDE model are computed using noise-robust gradient operators; (III) computed gradients are used to directly estimate parameters; (IV) an attention layer based on the rate of temperature change (RTC) is used to compute weighted (spatially averaged) parameter estimates; (V) based on a confidence threshold on the RTC-field, final parameters are computed.}
  \label{fig:architecture}\vspace{-0.5cm}
\end{figure*}

\section{Introduction}
\label{sec:intro}


Over the past decades, the advantages of robot-assisted surgery over conventional laparoscopy have become increasingly apparent \cite{Lanfranco2004}. As the robotic surgery performance demands grow, so does the need for a deeper understanding of the physical phenomena governing both the tools and the tissue treated, as well as the control laws for the precise attainment of the surgical objectives. To enable the latter autonomous control and decision algorithms, as well as improving the practicing physician's awareness when dealing with hard-to-observe tissue dynamics, it is imperative to obtain high-fidelity models of the tissue biomechanics as adapted on a patient-by-patient basis. In the case of \emph{energy-based surgical techniques}, which have garnered significant interest over the past decades in minimally invasive surgical procedures \cite{Palanker2008a}, access to high-fidelity tissue thermodynamic models is indispensable in predicting the tissue thermal response, and the resulting thermal secondary damage footprint \cite{El-Kebir2021b}. These high-fidelity thermodynamics models can be used in robot-assisted surgery, but also for the real-time monitoring and estimation of heat spread underneath the tissue surface when performing minimally invasive energy-based surgery, in which secondary tissue damage due to thermal shock is to be minimized, yet its effects are hard to observe during surgery \cite{Singh2020}. Naturally, such thermodynamics models must be fit to the actual physical process before they can be employed for simulation or controller synthesis. This process of \emph{model} or \emph{parameter estimation} is the focus of this work. In particular, we highlight the utility of our parameter estimation approach as it relates to electrosurgical processes \cite{El-Kebir2021c, El-Kebir2021d}.

Infrared (IR) thermography (IRT) allows one to measure surface temperature fields at high frequencies and resolutions, simply by capturing infrared radiation in the long-infrared band (9--14 $\mu$m) without any active external excitation \cite{Gaussorgues1994}. In \cite{El-Kebir2021c}, it was for the first time proposed to use non-collocated point-wise temperature feedback (as is seen in thermographic cameras) for control of electrosurgical processes. This idea was first put to practice in \cite{El-Kebir2021d}, showing that real-time feedback control based on thermography of the tissue is indeed possible. However, most of the control laws in use today employ some internal model, which is based on thermophysical parameters of the thermodynamic process to be controlled. For this reason, some form of thermodynamic modeling is required, often prior to deploying the controller. In addition, due to the impossibility of measuring the through-thickness tissue thermal response, accurate thermodynamics models are necessary to estimate and predict the inner-domain thermal response to surface and near-surface heat sources.


Modern approaches to thermodynamic modeling try to employ thermographic data, but often resort to computationally intensive optimization routines to fit the parameters based on this data. In recent years, genetic algorithms (GAs) have seen extensive use \cite{Melo2017}, mainly due to the fact that the Jacobian of the error system depending on the partial differential equation need not be known, and therefore is hard to obtain in practice. Even when using these Jacobian-free optimization algorithms, convergence to the global optimum may not be guaranteed, and there is a possibility that optimization results are contaminated by sensor noise. Finally, GAs require a prohibitive number of model error evaluations, especially when the parameter space is of high dimension, and the numerical PDE solver uses a fine spatiotemporal resolution. In \cite{Liu2016a}, a Laplace transform of the original PDE was used to fit on instead, allowing the use of an efficient least-squares algorithm. This is only possible when dealing with linear dynamics, and becomes analytically prohibitive when the problem has higher spatial dimension, and also poses questions regarding noise-robustness. In work applied to electrosurgical joining of tissue, the authors of \cite{Yang2018, Yang2020} have obtained estimates of the tissue's thermal conductivity based on optimization over the PDE solution using a gradient-free optimization algorithm; this precludes real-time application of their method. In addition, the parameter search space was limited to attain faster convergence; this does, however, introduce preconceived biases with respect to the thermal conductivity, which may not capture the true value due to physiological effects such as dehydration or electrolytic imbalance \cite{Duck1990}.

Given that there are many variations in the governing coefficients on a patient-by-patient basis \cite{Ahmed2005}, in order to conduct safe autonomous electrosurgery, it is required that a good estimate of the thermal properties be obtained in real-time, without the need for any invasive laboratory testing. Ideally, this learning procedure should be carried out for the entirety of an electrosurgical process, without disturbing the process itself, such that any changes in thermophysical parameters may be identified and acted upon.

When high-resolution spatiotemporal data of a thermodynamic process is available, one would prefer a direct extraction of the process dynamics, rather than an indirect PDE solution computation and parameter estimation. We propose, to the best of our knowledge, the first \emph{direct approach} in which the spatial and time derivatives of the thermographic readings are estimated, and a PDE model is directly fitted to the readings. Since this procedure is susceptible to high frequency sensor noise, we propose the use of noise-robust convolutional gradient operators. In addition, we introduce a novel weighting approach based on the \emph{rate of thermal change}, so as to favor the regions in observed thermal field with high thermal activity. The essence of our approach is to \emph{balance the underlying differential equations} based on the observed thermographic readings, rather than optimizing over the PDE solutions. We call our approach \emph{attention-based noise robust averaging} (ANRA). Our goal is to obtain model parameter estimates for use in apparent thermodynamics models based on the surface temperature dynamics. The inner-domain extensions based on the latter have been explored in \cite{Toivanen2014, El-Kebir2022c}, but ground truth model estimates remain hard to obtain in general, let alone in real-time. In spite of this limitation, in this work, we perform both verification of our method in simulation, as well as experimental validation on \emph{in vivo} porcine epidermis, with the nominal values (``ground truth") being adopted from the literature. The proposed model estimation architecture is described in Fig.~\ref{fig:architecture}.

This paper is organized as follows. In Sec.~\ref{sec:Preliminaries}, the model structure that will be considered is introduced, and the model identification problem is identified. Sec.~\ref{sec:Model Architecture} presents the model architecture that underpins our learning method. In addition, simulation results demonstrating the efficacy of our method compared to other possible approaches are shown. In Sec.~\ref{sec:Applications}, we apply our method to real-life thermographic data from \emph{in vivo} electrosurgery on porcine epidermal tissue. Conclusions and future research directions are presented in Sec.~\ref{sec:Conclusion}.




\begin{figure}[b]
\centering
\centerline{\includegraphics[width=\linewidth]{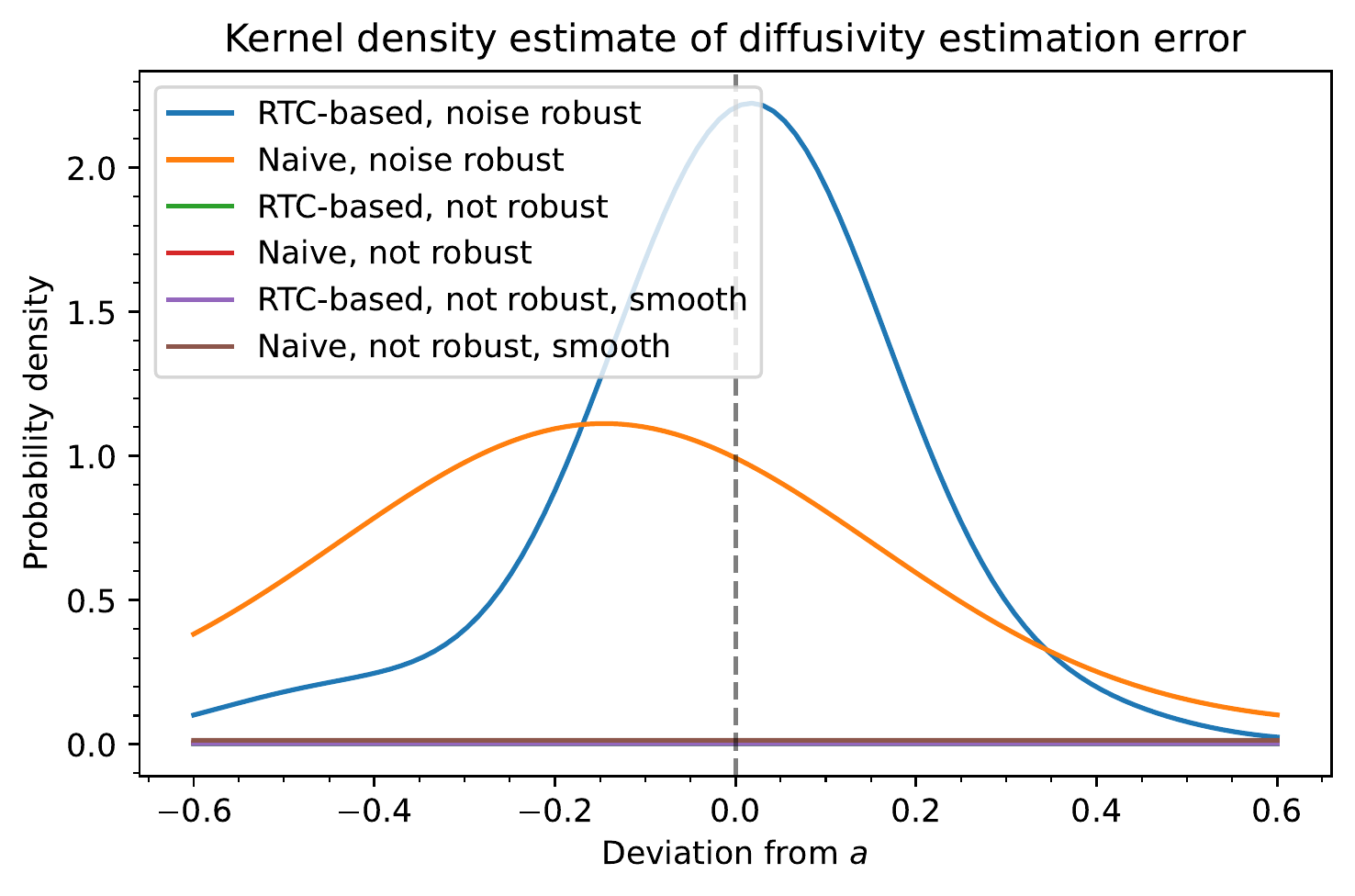}}
  \caption{Kernel density estimates for diffusivity estimate deviation in Problem~\ref{prob:FHT}. The sharp bell-shaped curve is obtained when applying ANRA; the bottom four lines in the legend have probability density close to zero in the plotted domain.}
  \label{fig:KDE FHT}
\end{figure}

\begin{figure*}[htb]
\begin{minipage}[b]{.3\linewidth}
  \centering
  \centerline{\includegraphics[width=\linewidth]{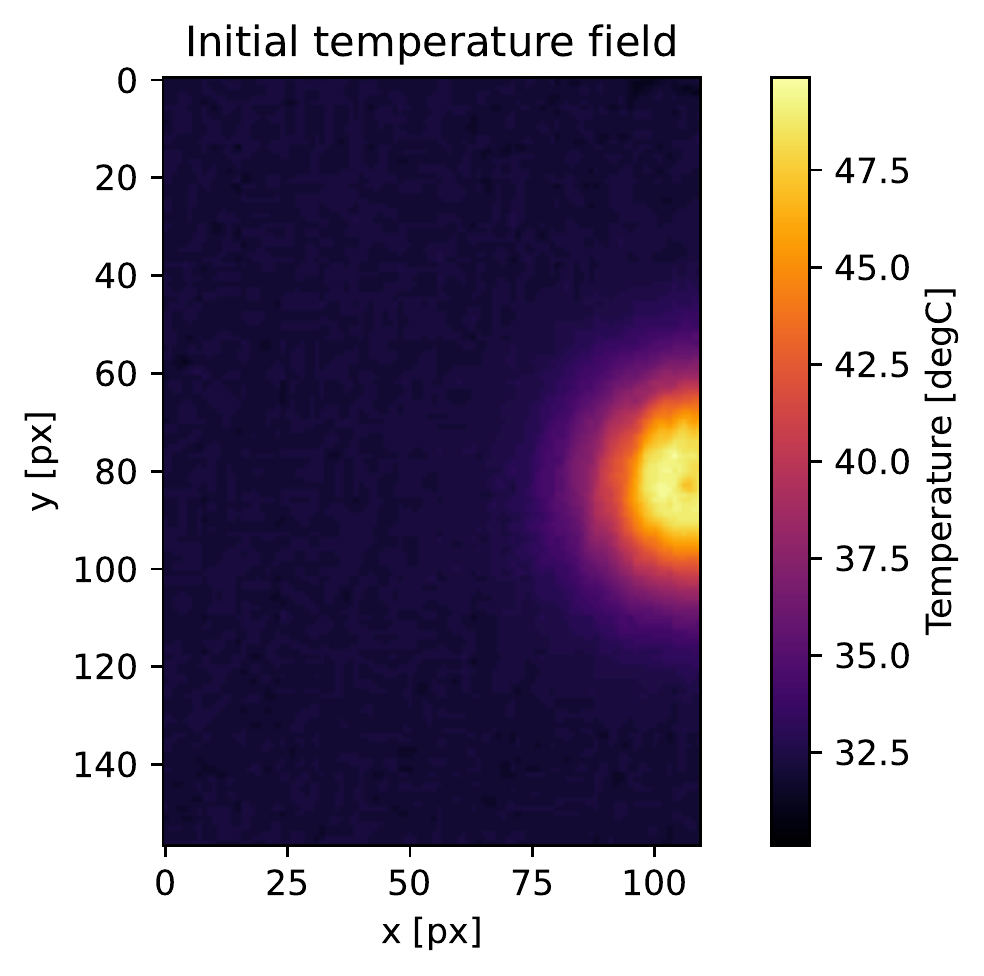}}
  \centerline{(a) Initial measured temperature field.}\medskip
\end{minipage}
\hfill
\begin{minipage}[b]{.3\linewidth}
  \centering
  \centerline{\includegraphics[width=\linewidth]{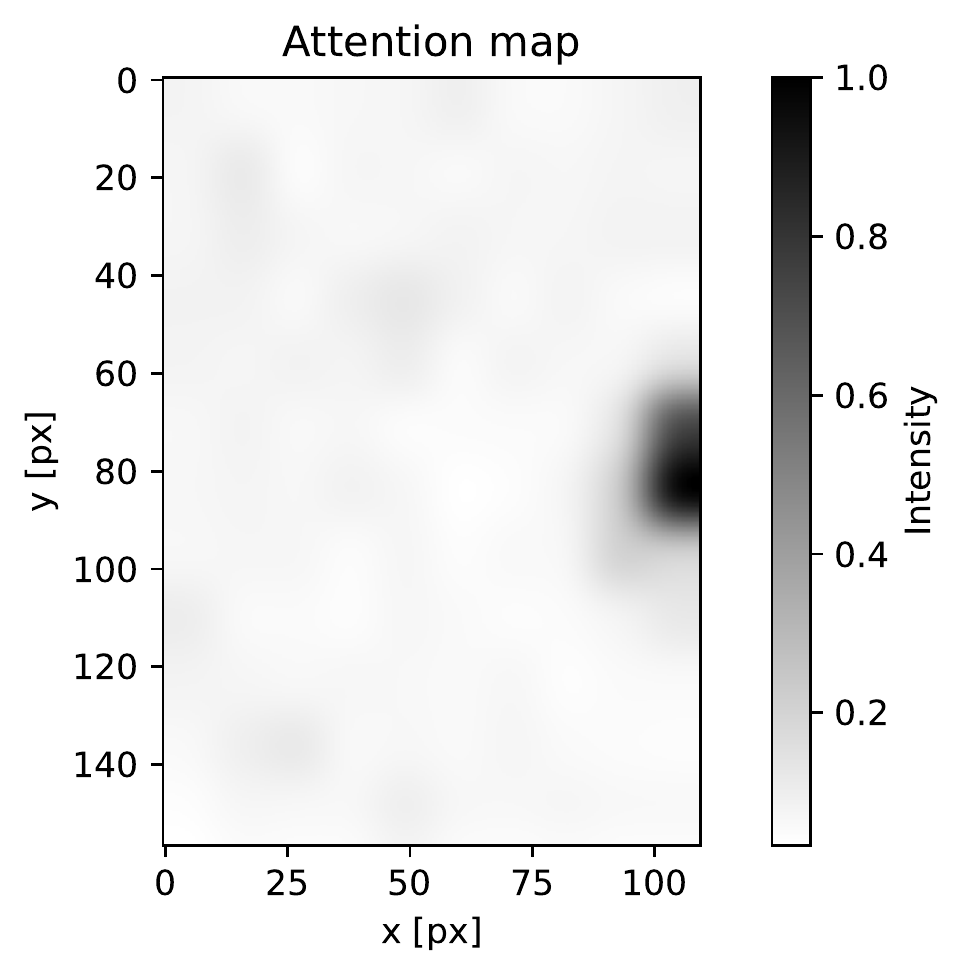}}
  \centerline{(b) RTC-based attention layer.}\medskip
\end{minipage}
\hfill
\begin{minipage}[b]{0.3\linewidth}
  \centering
  \centerline{\includegraphics[width=\linewidth]{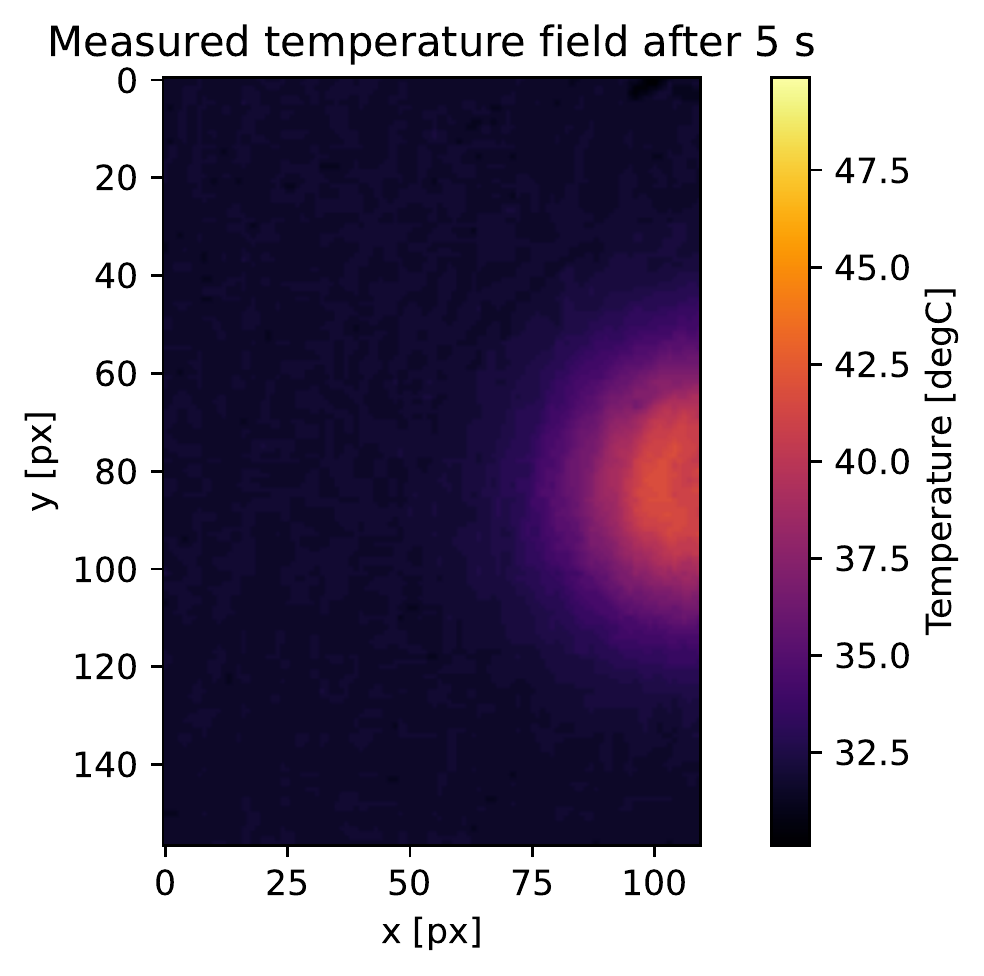}}
  \centerline{(c) Measured temperature field after 5 seconds.}\medskip
\end{minipage}
\hfill
\begin{minipage}[b]{.37\linewidth}
  \centering
  \centerline{\includegraphics[width=\linewidth]{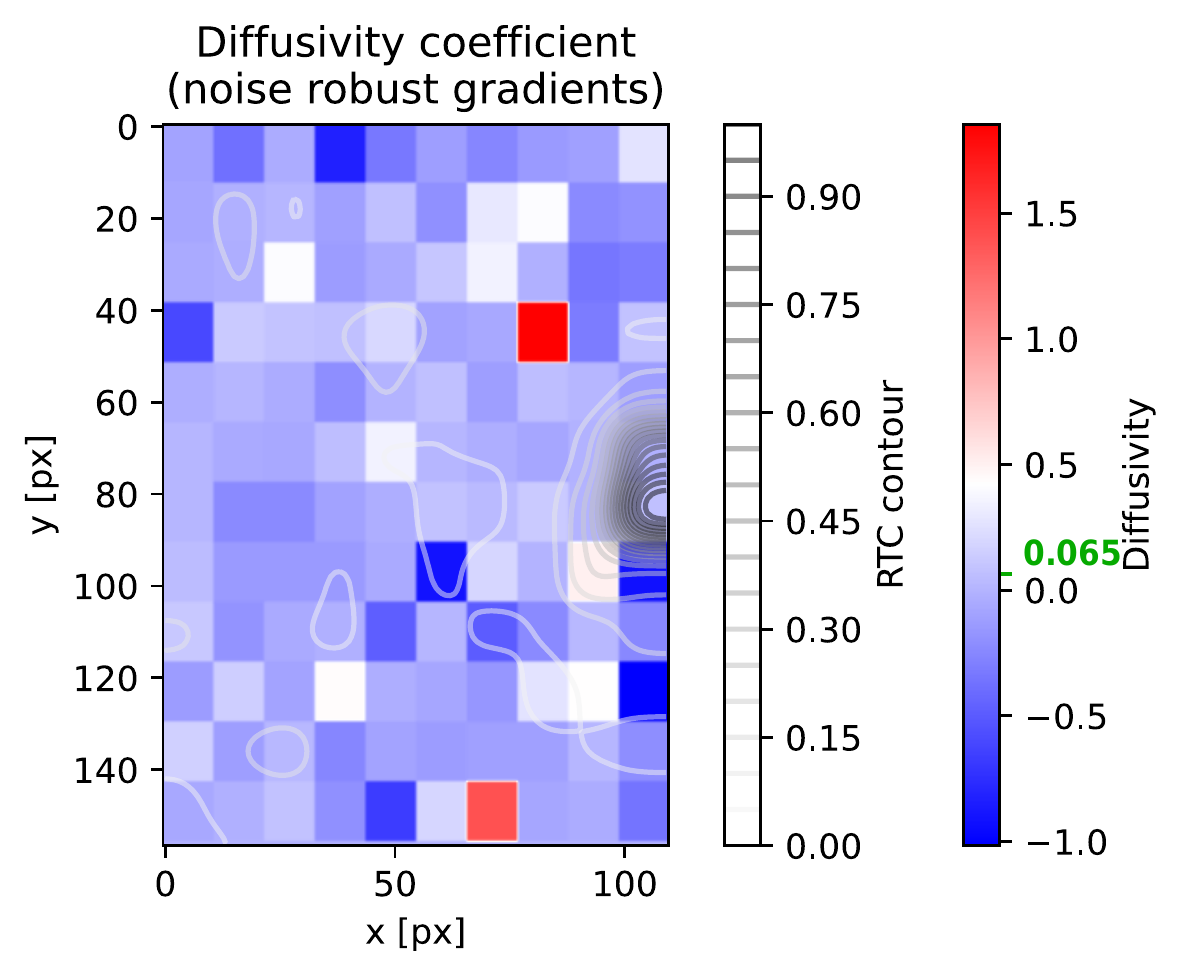}}
  \centerline{(d) Diffusivity estimate field.}\medskip
\end{minipage}
\hfill
\begin{minipage}[b]{.3\linewidth}
  \centering
  \centerline{\includegraphics[width=\linewidth]{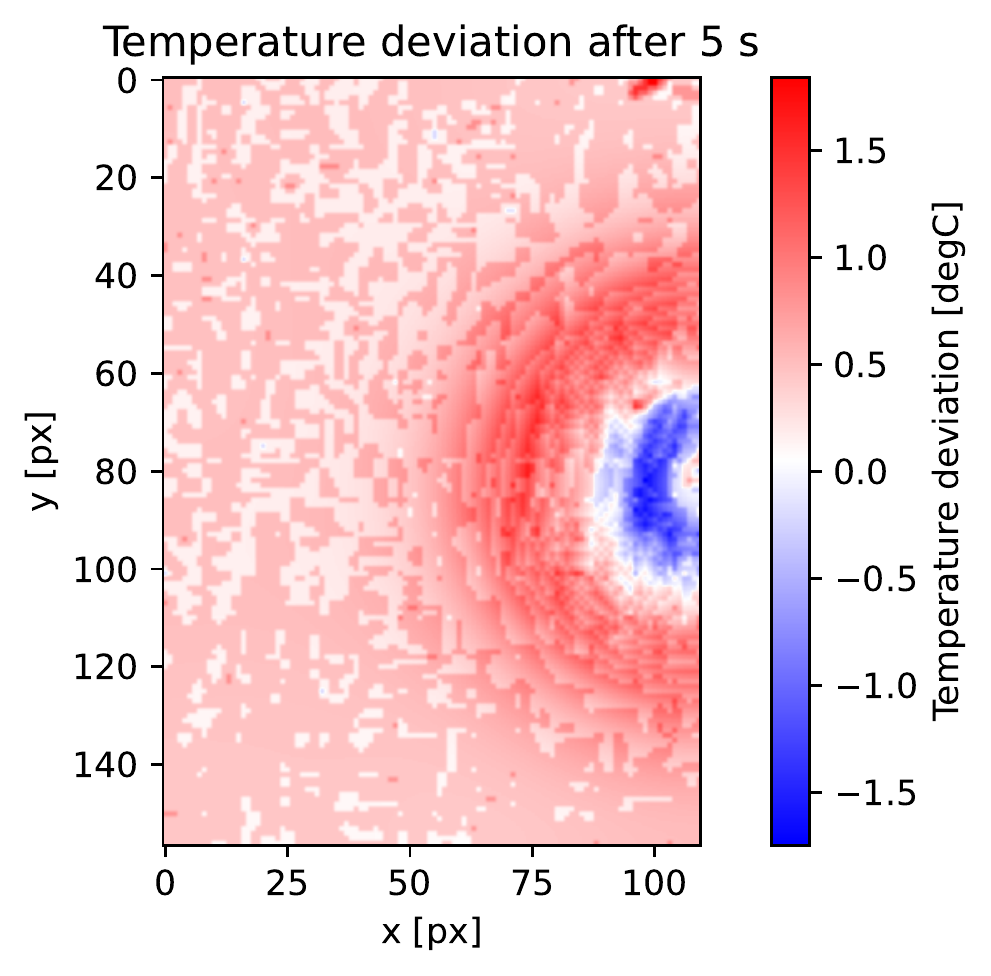}}
  \centerline{(e) Temp. field deviation after 5 sec.}\medskip
\end{minipage}
\hfill
\begin{minipage}[b]{0.3\linewidth}
  \centering
  \centerline{\includegraphics[width=\linewidth]{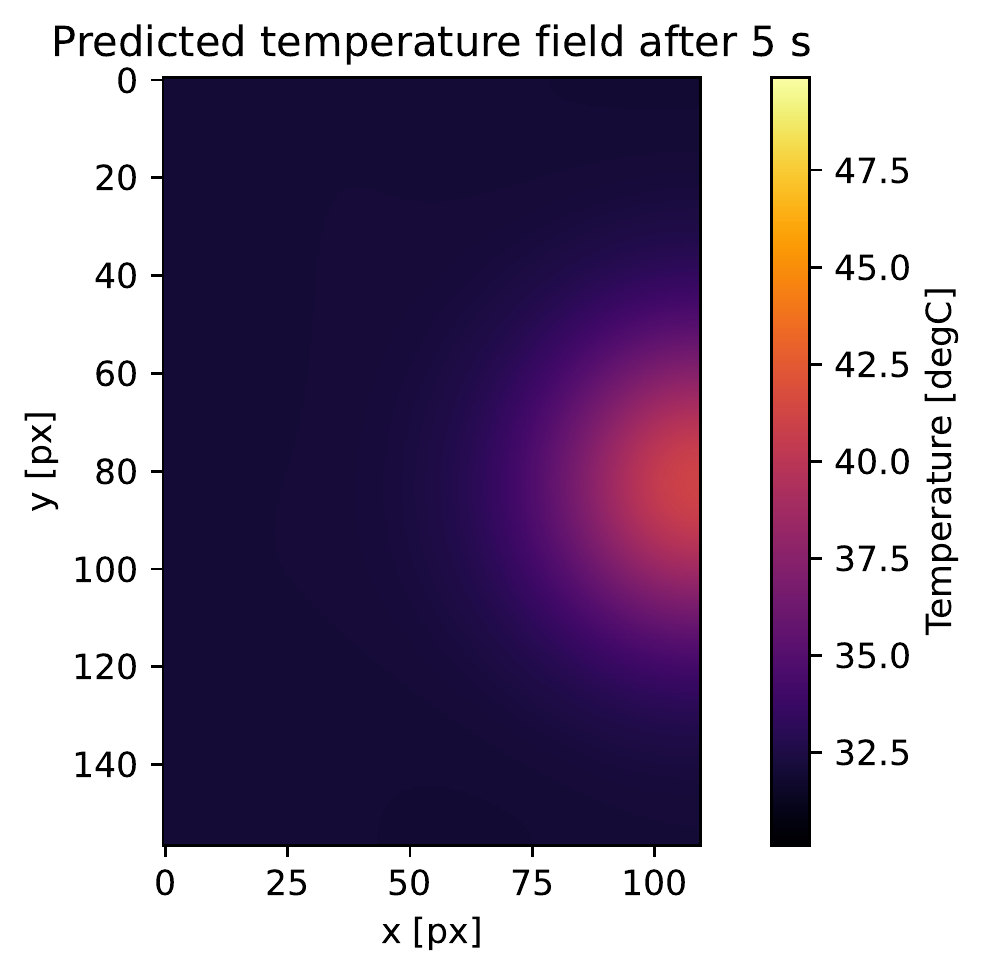}}
  \centerline{(f) Predicted temp. field after 5 seconds.}\medskip
\end{minipage}
\caption{Application of the windowed direct estimation approach to Problem~\ref{prob:FHT} (diffusivity estimation) on \emph{in vivo} porcine epidermis data. The measured and the predicted temperature fields after 5 seconds shown in the rightmost images are seen to be close to identical.}
\label{fig:skin FHT}
\vspace{-0.5cm}
\end{figure*}

\section{Preliminaries}
\label{sec:Preliminaries}

We consider the classical Fourier heat transfer equation (FHT) in two dimensions:
\begin{equation}\label{eq:FHT}
    \partial_t T = a (\partial_{11} + \partial_{22}) T.
\end{equation}

In this work, we aim to solve the following problem:

\begin{problem}[Diffusivity Fitting for FHT]\label{prob:FHT}
Given observer temperature fields $\{ \hat{T}[k] \}_{k=0}^N$ sampled at a period $\Delta t > 0$, find $\hat{a}$ (either a scalar, or a spatially varying field) in
\begin{equation}
    \partial_t \bar{T}_{\hat{a}} = \hat{a} (\partial_{11} + \partial_{22}) \bar{T}_{\hat{a}}, \quad \bar{T}_{\hat{a}}(0) = \hat{T}[0]
\end{equation}
such that the error
    $e_{\hat{a}} := \sum_{k = 0}^{N} \Vert \hat{T}[k] - \bar{T}_{\hat{a}} (k \Delta t) \Vert_2^2$
is minimized.
\end{problem}


Classically, Problem~\ref{prob:FHT} would be solved using some optimization routine, but this is untenable in real-time, since this requires repeated solving of the model PDE. Hence, we aim for the real-time-obtainable suboptimal solutions to both problems. A key standing requirement is therefore that there be \emph{no evaluation of any PDE solutions} in our method.

\section{Attention-based noise robust averaging}
\label{sec:Model Architecture}

The learning framework proposed includes three main components: (i) numerical gradient computation from thermal fields; (ii) attention layer definition; (iii) direct parameter estimation. The first component, the numerical gradient computation, is based on finite differencing and noise-robust filtering, as described next.

\subsection{Numerical Gradient Computation}

Since our learning architecture depends on direct computation of the spatial and temporal temperature gradients, we wish to suppress the effects of high frequency noise, which is observed both spatially and temporally in thermographic data. For this reason, a \emph{noise-robust filter}, i.e., one that subdues high frequency noise while leaving low frequency content unattenuated, is desirable \cite{Holoborodko2008}. Pavel Holoborodko introduced noise-robust gradient operators, in backward differencing form \cite{Holoborodko2008}, as well as linear filters for spatial gradients \cite{Holoborodko2009}; this family of operators is instrumental in our work.


\subsection{Attention Layer Definition}

In this work, we introduce a \emph{rate of thermal change} (RTC) field which indicates which areas of the temperature field exhibit most activity. For this reason, we propose to draw the attention of a learning algorithm to areas of high RTC, using a data-driven \emph{attention layer}.

This focus on the areas of high thermal activity serves a triple purpose: first, it actively incentivizes learning new dynamics, second, it draws attention away from the areas likely dominated by noise, and third, it maximizes the learning rate by focusing on the areas supporting the shortest learning time for capturing the parameters of interest.

Let $\hat{T}[k] \in \mathbb{R}^{N, M}$ be the observed temperature field at time index $k \in \mathbb{N}_0 := \{ 0 \} \cup \mathbb{N}$. We assume that the temperature fields are sampled at a fixed period $\Delta t > 0$. Let $\hat{\partial}_t$ be some finite-difference based time derivative operator, and let $\hat{\partial}_{ij} = \hat{\partial}_{j} \hat{\partial}_{i}$ be the product of two spatial differential operators, where $i, j \in \{1, 2\}$.

We define the rate of thermal change (RTC) as follows: \small
\begin{equation}\begin{split}
\text{RTC}_{i,j} [k] &\leftarrow \exp(|\hat{\partial}_t \hat{T}_{i,j}[k]|) \\
\text{RTC}_{i,j} [k] &\leftarrow \text{RTC}_{i,j} [k] - \min_{i,j} \text{RTC}_{i,j}[k] \\
\text{RTC}[k] &\leftarrow \text{RTC}[k]/ \left(\sum_{i,j} \text{RTC}_{i,j}[k] \right).
\end{split}\end{equation}\normalsize
The choice for this definition is as follows. We take the absolute value of the temperature gradient at each location and exponentiate it, so as to assign a larger positive value to a large thermal rate of change, this value scaling nonlinearly in the gradient. We then subtract the smallest element to get the minimum to be zero. Finally, we divide by the sum of all elements, so as to obtain a matrix whose elements sum to 1. This latter property is instrumental as shown next.

The RTC layer serves to disregard points likely to be noise-driven, and instead favors regions with high thermal activity. This gives rise to a weighted averaging step, as shown next.

\subsection{Direct Parameter Estimation}

In this work, we consider the Fourier heat transfer (FHT) equation of Problem~\ref{prob:FHT}. There are two main approaches to fitting the thermal diffusivity $a$ in \eqref{eq:FHT}; a global approach, as well as a windowed approach. 
Central to both ideas is the notion of the rate of temperature change (RTC) presented in the previous section.

We estimate $a$ in \eqref{eq:FHT} directly as follows:
\begin{equation}\label{eq:RTC diffusivity global}
    \hat{a}[k] = \sum_{i,j} \text{RTC}_{i,j}[k] \left\{ \hat{\partial}_t \hat{T}[k] / [(\hat{\partial}_{11} + \hat{\partial}_{22}) \hat{T}[k]] \right\}_{i,j}.
\end{equation}



We now proceed to show this approach in action on a simulated example. We also provide a performance comparison of the latter when instead the non-noise-robust gradient operators are used.

We take $a = 0.2$, and simulate \eqref{eq:FHT} based on a Gaussian initial temperature field with a peak temperature of 3.8 units, of diameter 50 pixels, similar to what one would capture on a thermographer (see, e.g., Fig.~\ref{fig:skin FHT}(a)). We have added a zero-mean Gaussian noise, with standard deviation 0.01 to all simulation results prior to parameter estimation. Our observation data spans 1 second, and is sampled at 27 Hz. The spatial resolution is 224 pixels squared, with a physical length of 10 by 10 units.


One can compare the performance of our method to approaches using classical image processing techniques. These include using non-noise-robust filters and smoothing. Both approaches attenuate the dynamics by amplifying noise or introducing artificial diffusion. The effect of these choices can clearly be seen in the kernel density estimate of the final thermal diffusivity error, as shown in Fig.~\ref{fig:KDE FHT}.

Our method produced an estimate of $\hat{a} = 0.189$ (recall that the ground truth is 0.2), while backward differencing combined with a Laplacian of Gaussian filter produced $\hat{a} = -7.9$, and adding additional Gaussian smoothing resulted in $\hat{a} = -41.9$, with the latter two estimates being physically infeasible.

\section{Application}
\label{sec:Applications}

We apply the technique presented to obtain the thermal diffusivity estimate of \emph{in vivo} porcine epidermis (skin tissue), after exposure to 10 seconds of 15 watts of pure cutting monopolar electrosurgical action using a Covidien Force FX-3 electrosurgical generator. We used a dual Optris Xi 400 microscopic thermographer setup, positioned about 11 centimeters from the tissue, and logging at 27 Hz.

%

Using our approach, we find a thermal diffusivity of $\hat{a} \approx 0.065$ mm\textsuperscript{2}/s (see Fig.~\ref{fig:skin FHT}(d)), which agrees very closely with data from the literature. Andrews \emph{et al.} \cite{Andrews2016} report a thermal diffusivity of $0.03 \pm 0.02$ mm\textsuperscript{2}/s, which closely aligns with our result. In Fig.~\ref{fig:skin FHT}(e), we show the temperature deviation between the observed and the predicted values (using the initial condition in Fig.~\ref{fig:skin FHT}(a) and $\hat{a}$ in \eqref{eq:FHT}). Figs.~\ref{fig:skin FHT}(c) and \ref{fig:skin FHT}(f) show that the parameter estimate obtained is very likely to be close to optimal for the choice of model. The RTC-based attention layer of Fig.~\ref{fig:skin FHT}(b) could also be used, e.g., for tissue damage classification.

\section{Conclusion}
\label{sec:Conclusion}

In this work, we have presented a novel approach to real-time learning of thermodynamics using thermographic data. Unlike past methods which require solution of complex PDE systems, our approach, \emph{attention-based noise robust averaging} (ANRA), \emph{operates directly on thermographic data and requires only the PDE structure to be given}. In addition, our method relies on thermographic readings, which are fully non-intrusive in nature. We have applied ANRA to a theoretical test case involving Fourier heat transfer, as well as the real-life data from an \emph{in vivo} porcine specimen subject to electrosurgical action. In both cases, our approach clearly outperforms naive real-time estimation methods, suggesting that ANRA presents a robust real-time alternative to offline optimization methods.

The future work will address the extension of this approach to other thermodynamic models, such as the reaction-diffusion equation, as well as possible applications of tissue classification based on the thermal response.

\bibliographystyle{IEEEtran}
\bibliography{tmi.bib}

\end{document}